\documentclass[letterpaper]{article} 
\usepackage{aaai19}  
\usepackage{times}  
\usepackage{helvet}  
\usepackage{courier}  
\usepackage{url}  
\usepackage{graphicx}  

\usepackage{subfigure} 
\usepackage{color}
\usepackage{amsmath}
\usepackage{amsfonts}
\usepackage{mathrsfs}
\usepackage{slashbox}
\usepackage{algorithm}
\usepackage{algorithmic}
\usepackage{wrapfig}
\usepackage{enumitem}
\usepackage{setspace}
\usepackage{wrapfig}
\usepackage{array}
\usepackage{multirow}
\usepackage{epsfig}
\usepackage{textcomp}

\usepackage[utf8]{inputenc} 
\usepackage[T1]{fontenc}    
\usepackage{booktabs}       
\usepackage{nicefrac}       
\usepackage{microtype}      
\usepackage{natbib}

\newcolumntype{M}[1]{>{\centering\arraybackslash}m{#1}}
\newcolumntype{P}[1]{>{\centering\arraybackslash}p{#1}}

\setlist[itemize]{leftmargin=5.mm}

\newcommand{\R}{\mathbb{R}}

\begin{document}
%
\title{Controllable Top-down Feature Transformer}
\author{Zhiwei Jia \and Haoshen Hong \and Siyang Wang \and Kwonjoon Lee \and Zhuowen Tu \\
 University of California, San Diego}

\maketitle

\begin{abstract}

We study the intrinsic transformation of feature maps across convolutional network layers with explicit top-down control. To this end, we develop top-down feature transformer (TFT), under controllable parameters, that are able to account for the hidden layer transformation while maintaining the overall consistency across layers. The learned generators capture the underlying feature transformation processes that are independent of particular training images. Our proposed TFT framework brings insights to and helps the understanding of, an important problem of studying the CNN internal feature representation and transformation under the top-down processes. In the case of spatial transformations, we demonstrate the significant advantage of TFT over existing data-driven approaches 
in building data-independent transformations. We also show that it can be adopted in other applications such as data augmentation and image style transfer.
\end{abstract}

\section{Introduction}

Over the past few years, deep neural networks \citep{krizhevsky2012imagenet,he2016deep} have led to tremendous performance improvement on large-scale image classification \citep{russakovsky2014imagenet} and other computer vision applications \citep{girshick2014rich,goodfellow2014generative,long2015fully,xie2015holistically,dosovitskiy2015flownet}. While convolutional neural networks (CNNs) have shown great promise in solving many challenging vision problems, there remain fundamental questions about the transparency of representations in current CNN architectures. While the explicit role of the top-down process is a critical issue in perception and cognition, it has received less attention within the current CNN literature.

Currently, both training and testing of CNNs is performed in a data-driven manner by passing convolved features from lower layers to the top layers. However, visual perception systems are shown to engage both bottom-up and top-down processes \citep{stroop1935studies,hill2007hollow}. A top-down process would allow explicit generation and inference of transformations and (high level) configuration changes of images that is otherwise not convenient in a bottom-up process. For example, suppose we wish to train a CNN classifier to detect the translation of an object in an image. A data-driven way to train this CNN would require generating thousands of samples by moving the object around in the image. However, a top-down model, if available, can directly detect translation using two parameters of the translation. Computational models realizing the bottom-up and top-down visual inference have been previously proposed \citep{marr1982vision,kersten2004object,tu2005image}. They, however, are not readily integrated into end-to-end deep learning frameworks.
Recurrent neural networks (RNN) \citep{RNN,LongShortMemory} have feedbacks recursively propagated between the output and input layers, but they miss explicit top-down generation.

Motivated by the recent development in CNNs \citep{krizhevsky2012imagenet} that learn effective hierarchical representations, the general pattern theory \citep{grenander1993general,yuille1992feature,zhu2007stochastic} that provides rigorous mathematical formulation for top-down generations, as well as findings from cognitive perception \citep{gregory1970intelligent,dodwell1983lie,gibson2002theory}, we seek  to build a top-down generator, {\em under controllable parameters}, that operates directly on the feature maps of the internal CNN layers to model and account for the underlying transformations.

In this paper, we pay particular attention to CNN features under top-down transformations. There often exist clear flow fields computed between the extracted features of the original image and those of the transformed images (after rotation, scaling, and translation) \citep{gallagher2015happened}. Such a pattern of consistent but nontrivial feature map deformations throughout the convolutional layers is a key topic to be studied and leveraged here. Our goal is to discover and model operations in CNNs that lead to non-linear activity of the resulting flow fields. Given a source image and a transformed image under translation, rotation, and scaling, the internal CNN feature maps across multiple-layers can be directly computed; feature transformers modeled using an aggregated convolution strategy are themselves learned to perform mappings that transfer the feature maps of source image to the target image across all the intermediate CNN layers. 

The training process is supervised since we generate transformed images using different parameters for translation, rotation, and scaling. Given that no supervision is needed to have the explicit correspondences at the feature map level and transforming the source images can be readily accomplished by using the explicit transformation parameters, obtaining the training data can be done effortlessly. The learned top-down feature transformer (TFT) however demonstrates great generalization capable of transforming images that are not seen during training --- a benefit of having a top-down generator that is not tied to specific images. TFT is therefore distinct from existing work \citep{dosovitskiy2015learning,reed2015deep,gardner2015deep} where transformations are learned with strong coupling to the training images that are hard to generalize to novel ones.
For example, TFT is used learned to perform three kinds of flow transformations (rotation, translation and scaling) with arbitrary transformation parameters. Moreover, it generalizes well to various datasets of different input dimensions, ranging from small patterns to natural images. We also demonstrate TFT on the artistic style transferring task.
\begin{figure*}[!htp]
  \centering
  \begin{tabular} {c}
  \includegraphics[height=0.55\textwidth]{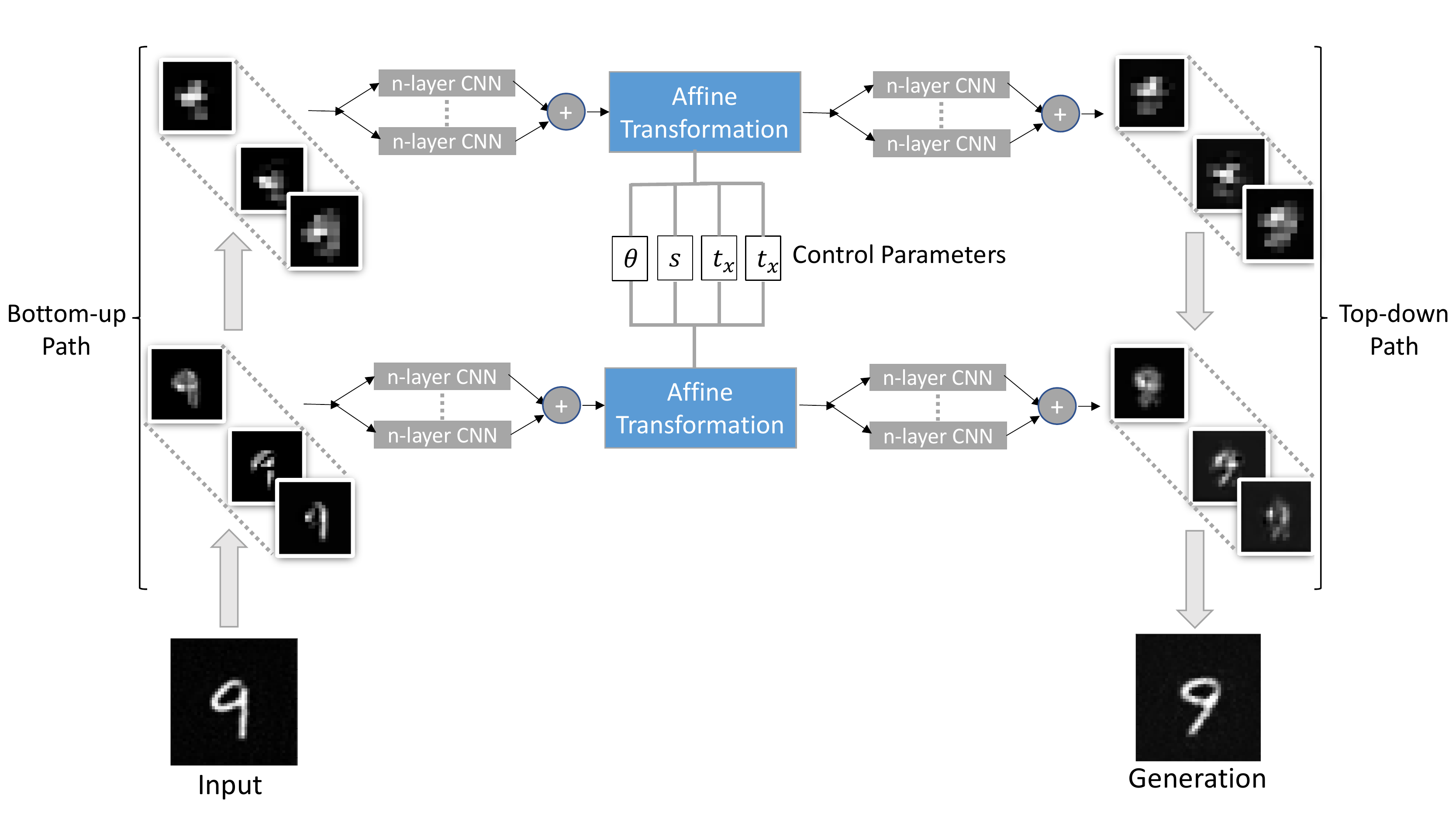}
  \end{tabular}
  \vspace{-3mm}
  \caption{A schematic illustration of our top-down feature transformer framework (TFT).}
\label{fig:overall}
\end{figure*} 
In the experiments, we train the proposed top-down feature transformer (TFT) on the MNIST dataset. We demonstrate that our TFT learns intrinsic flow transformations that are not tied to a specific dataset by ``inverting'' transformed CNN features of images taken from several non-MNIST datasets. Comparison with the competing transformer \citep{reed2015deep} shows the immediate benefit our approach. We further train TFT on images synthesized from PASCAL VOC 2012 \citep{pascal-voc-2012} and utilize it to fine-tune VGG-16 \citep{Simonyan14c} via network internal data augmentation to improve ImageNet \citep{ILSVRC15} classification performance. Lastly, we adapt our method to image style transfer to show that our method can be extended beyond spatial transformations. 

\section{Significance and Related Work}

We first discuss the significance of our proposed top-down feature transformer (TFT).

\noindent{\bf Why a top-down generator}? Top-down information can play a fundamental role in unraveling, understanding,
and enriching the great representation power of deep convolutional neural networks to bring more transparency and interoperability. The feature flow maps learned by TFT on novel images show promise for the direction of top-down learning.

\noindent{\bf Why transform features across CNN layers}? We are intrigued by the idea to understand how the CNN features change internally with respect to the spatial transformations. Capturing the generic feature flows under spatial transformations, our proposed TFT will aid in understanding the representation learned by a CNN in order to improve its robustness and to enrich its modeling and computing capabilities. This is not immediately available in the existing frameworks \citep{dosovitskiy2015learning,reed2015deep} where the models are heavily coupled with the specific training data. 

\noindent{\bf Why not Spatial Transform Networks (STN) \citep{jaderberg2015spatial}}? In \citep{jaderberg2015spatial}, a spatial transformer was developed to explicitly account for the spatial manipulation of the data. This differentiable module can be inserted into existing CNNs, giving neural networks the ability to actively transform the image. However, the main goal of \citep{jaderberg2015spatial} is to learn a differentiable transformation field through backpropagation to account for the spatial transformations using a localization network for classification to obtain the transformation parameters including rotation, scaling, translation, and sheering. In short, STN learns to perform spatial transformation, {\em without  controllable/user specified transformation parameters}, to match the output features whereas TFT studies the generator (with user specified transformation parameters) modeling the underlying changes to the feature maps in the result of spatial transformations. In the experiments section, we show that STN is indeed not designed for and not easily extendable for TFT’s tasks. Furthermore, we extend TFT to the style transfer problem to show the flexibility of our method.
In the experiments (section 5.1), we show a direct comparison to STN for modeling spatial transformations with controllable parameters.

\noindent{\bf Why not to perform spatial transformation directly}? The consistency and integrity of an image after a direct e.g. spatial transformation on the input image space cannot be all maintained. For example rotating or zooming out a certain part will create ghost regions/holes that need to be filled up. Performing learning-based feature transformation alleviates the problem of creating artifacts due to image transformation, as shown in Figure \ref{fig:natural_image}. In addition, TFT is not limited to just spatial transformation and it can be applied to many other tasks where feature transformation is needed such as online data augmentation, transfer learning, and image transformation. In Figure \ref{fig:style}, we show some results of TFT being applied to image style transfer. Moreover, we focus on the intrinsic feature change after top-down transformation, which is an important step towards understanding the transparency and unraveling the interpretability of CNN.

Next, we discuss related work.

\noindent{\em 1. Deep image analogy}. The deep visual analogy-making work \citep{reed2015deep} shows impressive results to learn to transform and compose images. However, \citep{reed2015deep,gardner2015deep} build heavily on an encoder-decoder strategy that is strongly tied to the training images; it learns to output results in the image space without modeling the internal feature maps. Applying learned transformer \citep{reed2015deep} to novel images therefore leads to unsatisfactory results, as shown in the experiment section. Instead, our approach studies flow transformations on CNN features and generalizes well. 

\noindent{\em 2. Learning image transformations}. Learning to transform images has been a quite active research area recently. Existing methods that target on building transformation generators \citep{jaderberg2015spatial,lin2016inverse,gregor2015draw,kulkarni2015deep,wu2017recursive} train CNN/RNN to perform transformation on the the output feature or the image space, which is different from our goal of studying the intrinsic feature transformations under top-down process within the CNN networks. The key difference between TFT and STN \citep{jaderberg2015spatial} has been discussed previously.

\noindent{\em 3. Feature transformation with SIFT-flow \citep{gallagher2015happened}}.
An earlier attempt \citep{gallagher2015happened} 
has been made to study the feature layer deformation under explicit transformations based on flows computed from the SIFT-flow method \citep{liu2011sift}. Although the work \citep{gallagher2015happened} has a similar big picture idea to our work here, but it is preliminary and builds transformation solely based on the SIFT-flow estimation \citep{liu2011sift}. It is therefore limited in several aspects: (1) only able to morph the features but cannot change the values; (2) may fail if SIFT-flow does not provide reliable estimation; (3) hard to generalize to arbitrary translation, rotation, and scaling. It relies heavily on the carefully chosen parameters that do not work well under general situations. Our approach has a better learning capability than that of \citep{gallagher2015happened}.

\noindent{\em 4. Generative Models}. We also discuss the existing literature in generative modeling. A family of mathematically well defined generators are defined in \citep{grenander1993general} as the general pattern theory. Its algorithmic implementation however still needs a great deal of further development. Methods \citep{blake1993active,cootes2001active,wu2010learning,zhu2007stochastic} developed prior to the deep learning era are inspiring, but they have limited modeling capabilities. Deep belief net (DBN) \citep{Hinton06} and generative adversarial networks (GAN) \citep{goodfellow2014generative} do not study the explicit top-down generator for the image transformation. Other generators that perform feed-forward mapping \citep{dosovitskiy2015learning,wu2016single,zhang2017growing} have transformations as input parameters. The process in \citep{dosovitskiy2015learning} maps directly from the input parameter space to the output image space; it cannot be applied to generate novel categories and does not study the intrinsic transformation.

\noindent{\em 5. Flow Estimation}. Existing works in flow estimation \citep{liu2011sift,dosovitskiy2015flownet} are used to perform flow estimation, not as a generator for feature transformations. Our frameworks examine the transformations in CNN features that can be applied to generate novel images from various datasets and to perform data augmentation in a network internal fashion.

\section{Top-down Feature Transformer}

\subsection{Architecture} 
The network comprises three layers: an aggregated feature transformation layer, an affine layer performing spatial transformation, and another aggregated feature transformation layer (which does not share weights with previous layers). Consider the feature map $f_x \in \R^{n\times n\times m}$ of a convolutional layer with $m$ channels from a CNN that is pretrained for a discriminative task (e.g., image classification) by feeding an image $x$. The aggregated feature transformation layer is given as: 
\[\mathcal{F}(f_x) =  w_0 f_x + \sum\limits_{i=1}^N  w_i{T}_i(f_x)\]
where $w_i$'s are parameters governing the weight of each branch, $N$ is an integer referred as the number of transformation functions for aggregation. Each transformation function $\mathcal{T}_i(\cdot)$ is defined as a convolutional network, where each layer has a convolution operation followed by the rectified linear (ReLU) activation. 

The transformed feature maps are then fed into an affine layer that applies spatial transformations, including translation, rotation and scaling, to each channel individually (with bilinear interpolations). The spatial transformations are modeled by the product of three transformation matrices:

$M_{rot}(\theta) = 
\begin{bmatrix} 
\cos \theta & -\sin \theta & 0 \\ \sin \theta & \cos \theta & 0 \\ 0 & 0 & 1
\end{bmatrix}$,
$M_{scale}(s) = 
\begin{bmatrix} 
s & 0 & 0 \\ 0 & s & 0 \\ 0 & 0 & 1
\end{bmatrix}$, and 
$M_{tran}(t_x, t_y) = 
\begin{bmatrix} 
1 & 0 & t_x \\ 0 & 1 & t_y \\ 0 & 0 & 1
\end{bmatrix}$
where $\theta$, $s$ and $(t_x, t_y)$ are top-down controls of rotation, scaling and translation, respectively. 
The results are further transformed by another aggregated residual layer to generate the output CNN features. Figure \ref{fig:overall} illustrates the architecture. We build the TFT for each convolutional layer in the pre-trained CNN. For clarity, we denote the collection of the overall transformation parameters as $\Theta = (\theta, s, t_x, t_y)$.

In the case of modeling spatial transformation, we manually set $w_0$ and half of the $w_i$ to $1$ and set the other half to $-1$. This configuration enables our method to model both the emerging patterns and the disappearing ones in CNN feature maps when spatial transformations are applied to the input images. We use differnt $w_i$ for the task of image style transfer, as discussed below.

\subsection{Model Training}
We train our proposed networks in a supervised manner by minimizing the average Euclidean distance $\mathcal{E} = \sum_{x, \Theta} {E_\Theta(x)}$ between the generated feature maps and the ground truth feature maps for any image $x$ and any transformation parametrized by $\Theta$ in the training set. The ground truth features maps can be collected automatically from CNN features of input images that are under the corresponding spatial transformations. In specific, for an image $x$ and its transformed version $\hat{x}$ under the transformation parametrized by $\Theta$, $E_\Theta(x)$ is given as $E_\Theta(x) = \left \| F_\Theta(f_x) - f_{\hat{x}} \right \|_2^2$
where $F_\Theta(f_x)$ is the generated feature maps by our TFT and $f_{\hat{x}}$ is the ground truth feature maps from the input image $\hat{x}$.

\subsection{Generating New Images}
Upon obtaining the transformed feature maps $\hat{f_x} = F_\Theta(f_x)$ for some transformation parametrized by $\Theta$,  we can generate images by ``inverting'' them in CNNs, similar to the process in \citep{gatys2015neural}, i.e., via back-propagation. We use transformed featrues from a set of layers altogether to generate images, with a TFT trained for each layer. One resulting benefit is better quality of generated images.

In practice, given a set of CNN layers $\{f_x(i)\}$, where $i$ represents the $i^{th}$ convolutional layer, we generate images by minimizing the combined representation loss $E(x_{new}) = \sum_i \alpha_i  \left \| f_{x_{new}}(i) - \hat{f}_x(i) \right \|_2^2 + \beta \cdot R_{TV}$ 
where $\alpha_i$ and $\beta$ are hyperparameters, $f_x(i)$ represents the feature maps from the $i^{th}$ CNN layer of the input image $x$. Similar to that of \citep{mahendran2014understanding}, we also add a regularization term $R_{TV} = \sum_{i, j} (x_{i+1,j}-x_{i,j})^2+(x_{i,j+1}-x_{i,j})^2$.

\subsection{Network Internal Data Augmentation}
As suggested in \citep{gallagher2015happened}, the learned generators can be applied to perform data augmentation inside the CNNs. Instead of feeding in newly generated images for CNN training, we can directly perform internal data augmentation in an online manner by applying learned TFT to transform the CNN features. In our experiments, we perform fine-tuning on CNN via TFT trained upon the same CNN; by this means, the proposed TFT that learns to capture CNN feature flow under spatial transformation can be used to improve CNN's invariance to such spatial transformations. We examine this in the experiments section.

\subsection{Flow Field Calculation}
A feature map with just one black pixel and all other pixels white, referred as a unit feature map, is fed into	the	feature	transformation model. The location of the black	pixel is the starting point of a flow. The center of mass of the output	feature	map	is calculated as the	end	point of the flow. Evenly spaced starting points are selected to calculate flows across	a feature map space to generate	a flow map of a	certain transformation under a feature transformation model. Our TFT is shown to learn clear flow fields while achieving non-linear transformations across CNN channels (Figure \ref{fig:comparison}).

\subsection{Image Style Transfer}
We adapt our proposed TFT to the task of image style transfer, where different kinds of styles are the new top-down control. In specific, we always feed an identity operation to the affine transformation layer as style transfer does not involve global spatial operations; we turn $w_i$, the weight associated with each feature transformation function, to values generated from style variable $c$ (one-hot encoded) via a two-layer fully-connected regression network. We find out that simply setting $w_0$ to 0 (discarding the residual connection) helps to make training more stable, likely due to the task of image style transfer. This experiment aims to show that TFT is easily extendable beyond spatial transformations, and we choose the total number of styles to be 10.

\section{Top-down Generator Design}

Our proposed TFT utilizes a combination of CNN-based feature transformations and explicit spatial transformations across all feature channels. As a result, TFT is particularly good at capturing the underlying flow transformation of the CNN feature maps. Our ablation studies indicate that removing any of the modules will result in poor performance of this task. Our method generalizes well to several datasets without compromising the promise of clear feature flow fields. The top-down approach with the use of controllable parameters makes our model very data-efficient compared to existing methods, as demonstrated below in the experiment section. 

The top-down approach remains effective and efficient in modeling underlying feature transformations even when we extend our method beyond spatial transformations by adapting the model, e.g., to the task of image style transfer. Specifically in our design, image styles as the top-down information directly controls weights associated with each of the feature transformation functions. These functions serve as basis in the image style space. It turns out that only a few image data are adequate for our method to learn and generate relatively good transformed images.

\section{Experiments}
\label{others}

We perform three experiments. The first two focus on modeling spatial transformations; the last one focus on image style transformations. 

In the first experiment, we train our top-down feature transformer (TFT) on the MNIST dataset and evaluate the learned generator on images from the notMNIST dataset. Under all three studied transformations, namely rotation, translation and scaling, we generate new images from transformed feature maps by applying TFT. We compare our results to those of \citep{reed2015deep} (DVAM) on MNIST and notMNIST dataset. We achieve better results both graphically and numerically. Our framework is able to perform out-of-bound transformations, i.e., transformations with arbitrary parameters that are out of the range in the training process. Our model can generalize well to new datasets as the learned flow transformations are generic and is much more data-efficient. Moreover, as TFT is not tied to fixed size of the input CNN features, it can perform feature transformations for arbitrary size of the input images, while \citep{reed2015deep} cannot. We also show that STN \citep{jaderberg2015spatial} cannot be adapted to explicitly taking user-specified parameters to generate images.

In the second and third experiments, we utilize TFT to perform network internal data augmentation and image style transfer, respectively.

\begin{figure}[!ht]
\centering
\vspace{-2mm}
\scalebox{0.57}{
\begin{tabular}{m{1cm}m{1.3cm}m{1.3cm}m{1.3cm}m{1.3cm}m{1.3cm}m{1.3cm}m{1.3cm}m{1.3cm}m{1.3cm}m{1.3cm}}
    \toprule
    \centering
     & Original Image & Rotation (30\textdegree) & Rotation (60\textdegree) & Rotation (90\textdegree) & Scaling (x0.9) &  Scaling (x1.1) & Trans-lation & Combi-nation\\ 
    \midrule
    DVAM & \raisebox{-\totalheight}{\includegraphics[width=1cm]{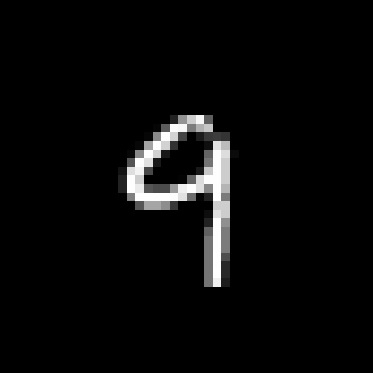}} & \raisebox{-\totalheight}{\includegraphics[width=1cm]{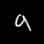}} & \raisebox{-\totalheight}{\includegraphics[width=1cm]{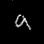}} & \raisebox{-\totalheight}{\includegraphics[width=1cm]{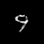}} & \raisebox{-\totalheight}{\includegraphics[width=1cm]{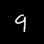}} & \raisebox{-\totalheight}{\includegraphics[width=1cm]{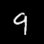}}& \raisebox{-\totalheight}{\includegraphics[width=1cm]{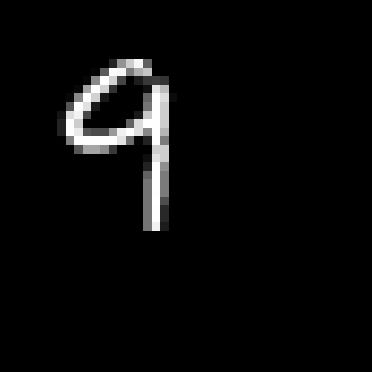}} & \raisebox{-\totalheight}{\includegraphics[width=1cm]{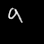}}\\
      Ours & \raisebox{-\totalheight}{\includegraphics[width=1cm]{MNIST-honglak/original.png}} & \raisebox{-\totalheight}{\includegraphics[height=1cm]{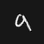}} & \raisebox{-\totalheight}{\includegraphics[height=1cm]{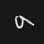}} & \raisebox{-\totalheight}{\includegraphics[height=1cm]{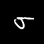}} & \raisebox{-\totalheight}{\includegraphics[height=1cm]{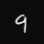}} & \raisebox{-\totalheight}{\includegraphics[height=1cm]{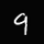}} & \raisebox{-\totalheight}{\includegraphics[height=1cm]{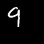}} & \raisebox{-\totalheight}{\includegraphics[height=1cm]{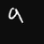}} \\  
    \midrule
    DVAM & \raisebox{-\totalheight}{\includegraphics[width=1cm]{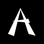}} & \raisebox{-\totalheight}{\includegraphics[width=1cm]{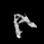}} & \raisebox{-\totalheight}{\includegraphics[width=1cm]{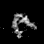}} & \raisebox{-\totalheight}{\includegraphics[width=1cm]{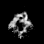}} & \raisebox{-\totalheight}{\includegraphics[width=1cm]{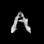}} & \raisebox{-\totalheight}{\includegraphics[width=1cm]{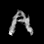}}& \raisebox{-\totalheight}{\includegraphics[width=1cm]{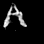}} & \raisebox{-\totalheight}{\includegraphics[width=1cm]{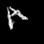}}\\
      Ours & \raisebox{-\totalheight}{\includegraphics[width=1cm]{notMNIST-honglak/original.png}} & \raisebox{-\totalheight}{\includegraphics[height=1cm]{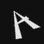}} & \raisebox{-\totalheight}{\includegraphics[height=1cm]{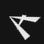}} & \raisebox{-\totalheight}{\includegraphics[height=1cm]{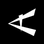}} & \raisebox{-\totalheight}{\includegraphics[height=1cm]{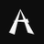}} & \raisebox{-\totalheight}{\includegraphics[height=1cm]{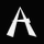}} & \raisebox{-\totalheight}{\includegraphics[height=1cm]{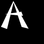}} & \raisebox{-\totalheight}{\includegraphics[height=1cm]{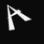}} \\  
    \bottomrule   
 \end{tabular}
} 
\caption{\footnotesize Comparison of transformations on the MNIST and notMNIST dataset. Rotations of $60^{\circ}$ and $90^{\circ}$ is beyond training settings (from $-30^{\circ}$ to $30^{\circ}$). Images generated by DVAM on notMNIST are vague and even lose the pattern when transformation parameters are out-of-bound. Images generated by our model show clear pattern of transformation.
}
\label{fig:comparison}
\end{figure}

\subsection{Training and Evaluation on MNIST}

We resize each image of dimension 28 x 28 in the MNIST dataset to 44 x 44 by zero-padding the original images so that each image has enough space to perform translation and scaling. For data in the training set, we perform a combination of all three studied transformations to the input images and output the CNN feature maps across different convolutional layers. Specifically, for translation, we perform two dimensional shifting of the input images, with each axis ranging from +7 to -7 pixels, i.e., 225 combinations. For rotation, we perform with 13 different angles, i.e., rotating the input images by \(5^{\circ}\), \(10^{\circ}\), \(15^{\circ}\), \(20^{\circ}\), \(25^{\circ}\), \(30^{\circ}\) clockwise and counterclockwise as well as \(0^{\circ}\). For scaling, we choose three factors: 0.9, 1.0 and 1.1 (1.0 indicates no scaling). These four parameters are the top-down transformation parameters in the training set. We form 3-tuples \((f_{x_{ori}}, f_{x_{\Theta}}, \Theta)\), where \(f_{x_{ori}}\) is the feature maps of the original images, \(f_{x_{\Theta}}\) is the feature maps of the corresponding images after performing spatial transformation parameterized by $\Theta$. Our training set is a collection of these 3-tuples. In practice, we apply the combinations of three transformations with randomly generated parameters from the range specified above.

In our experiments, we use a traditional CNN (with 3 convolutional layers, each is conv+relu+max-pooling and 2 fully connected layers) pre-trained on MNIST. We train three TFTs for the three convolutional layers, respectively, and use all these conv layers to generate new images. For each TFT, we set the depth of each transformation function as 2, the number of filters of the intermediate layer as 32, the filter size as 5x5, and the number of branhces $N$ as 8. 

The learning process of our TFT is supervised and the objective function is simply the Euclidean distance between the generated feature maps and the target ones, as mentioned in section 3.2. We train our networks by 200k steps with a \(L_2\) regularizer of coefficient 0.0001. We use the ADAM optimizer \citep{kingma2014adam} with learning rate of 0.0001. We set the batch size to be 128. 

We also train the networks proposed in \citep{reed2015deep}, namely DVAM in the same settings by using the same training set. We compare our graphical results of generated images from MNIST to those of DVAM in Figure \ref{fig:comparison}. When evaluated on MNIST, our approach outperforms theirs in terms of both feature flow representation and out-of-bound transformations. To further demonstrate our approach's data efficiency and capability of capturing the underlying feature transformation, we train another DVAM, denoted by DVAM+, in the same settings except (1) we double the rotation parameters in the training data (interval length becomes 2.5 degrees from five degrees on MNIST and (2) triple the number of epochs for training. The out-of-bound transformation of DVAM+ on MNIST (rotation in 60 and 90 degrees) is improved (mSPE, i.e., mean squared pixel error = 0.04438) but still much worse than that of TFT (mSPE = 0.000671). The cross-dataset performance on DVAM for rotation in general (30, 60 and 90 degrees) even drops, as reported in Table \ref{table:numerical_notmnist}.

Moreover, we train a STN with the same CNN as above and insert a spatial transformer (same localization net as in ST-CNN-multi in the original paper) before each convolutional layer. On MNIST, we evaluate transformed feature maps by STN by manually setting its parameters in each of the STs. It fails to model well. For example, mSPE is 0.1052 for rotation, much worse than TFT’s 0.000671. STN is indeed not designed for and not easily applicable to TFT’s tasks.

\begin{figure}[!h]
\centering
\vspace{-2mm}
\scalebox{1.8}{
\hspace{-2mm} \begin{tabular}{c}
  \centering
\includegraphics[width=0.5\columnwidth]{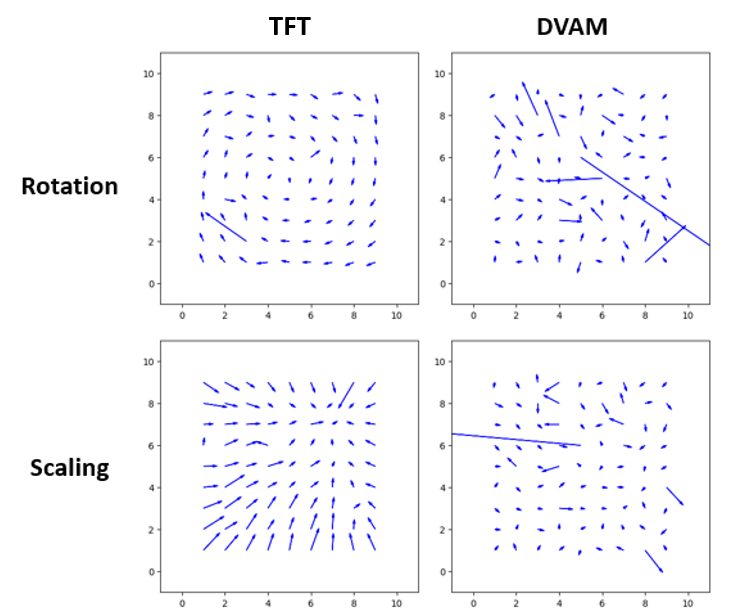}
\end{tabular}
}
\caption{\footnotesize Feature flows of TFT and DVAM \citep{reed2015deep}.
Flow generated by our model has clear pattern while the flow generated by DVAM does not.
}
\label{fig:comparison_flow}
\end{figure}

\vspace{-1mm}
\subsubsection{Learned Feature Flow}
\vspace{-0mm}
One critical advantage of our top-down feature transformer is its clear representation of learned feature flow. By using the method described in section 3.5, we compute feature flow fields resulted from our proposed TFT learned from the CNN pre-trained on MNIST. As a comparison, we perform the counterpart from DVAM in \citep{reed2015deep}, which has the similar CNN dimensions of the encoder-decoder structure.  We perform this flow experiment on MNIST images with rotation and scaling. The learned flow transformations in our model has clearer rotation and scaling deformation patterns than that of \citep{reed2015deep}, as in Figure \ref{fig:comparison_flow}.

\subsubsection{Out-of-bound Transformations}
Another advantage of our model is its robustness to out-of-bound parameters. We train our model on MNIST with rotation parameters ranging from $-30^{\circ}$ to $30^{\circ}$, yet we test the rotation transformation with $60^{\circ}$ and $90^{\circ}$. In comparison, we also perform this experiment on \citep{reed2015deep}. We can see that our model succeeds on the out-of-bound transformations while the model in \citep{reed2015deep} fails as illustrated in 3rd and 4th columns of Figure \ref{fig:comparison}.

\begin{figure}
\vspace{-6mm}
  \centering
  \scalebox{0.65}{
  \begin{tabular}{c||ccc||cc||c||c}
    \toprule
    Input & \multicolumn{3}{c}{\scriptsize Rotation} & \multicolumn{2}{c}{\scriptsize Scaling} & {\scriptsize Translation} & {\scriptsize Compositional} \\ 
    \midrule
      & 30$^o$ & 60$^o$ & 90$^o$ & 0.9 & 1.1 & varied & varied \\  
    \bottomrule
      \raisebox{-\totalheight}{\includegraphics[width=1cm]{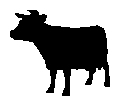}} & \raisebox{-\totalheight}{\includegraphics[width=1cm]{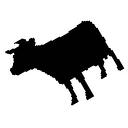}} & \raisebox{-\totalheight}{\includegraphics[width=1cm]{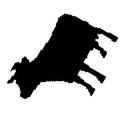}} & \raisebox{-\totalheight}{\includegraphics[width=1cm]{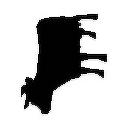}} & \raisebox{-\totalheight}{\includegraphics[width=1cm]{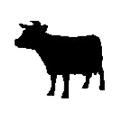}} & \raisebox{-\totalheight}{\includegraphics[width=1cm]{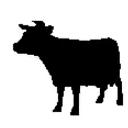}} & \raisebox{-\totalheight}{\includegraphics[width=1cm]{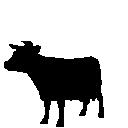}} & \raisebox{-\totalheight}{\includegraphics[width=1cm]{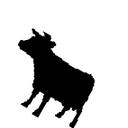}}    \\ 
      \raisebox{-\totalheight}{\includegraphics[width=1cm]{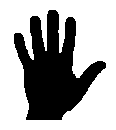}} & \raisebox{-\totalheight}{\includegraphics[width=1cm]{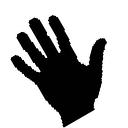}} & \raisebox{-\totalheight}{\includegraphics[width=1cm]{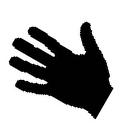}} & \raisebox{-\totalheight}{\includegraphics[width=1cm]{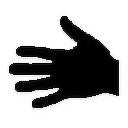}} & \raisebox{-\totalheight}{\includegraphics[width=1cm]{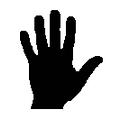}} & \raisebox{-\totalheight}{\includegraphics[width=1cm]{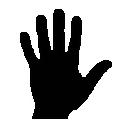}} & \raisebox{-\totalheight}{\includegraphics[width=1cm]{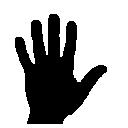}} & \raisebox{-\totalheight}{\includegraphics[width=1cm]{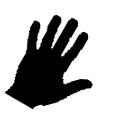}}    \\ 
  
  \bottomrule
      \raisebox{-\totalheight}{\includegraphics[width=1cm]{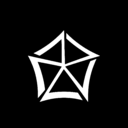}} & \raisebox{-\totalheight}{\includegraphics[width=1cm]{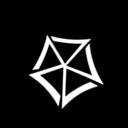}} & \raisebox{-\totalheight}{\includegraphics[width=1cm]{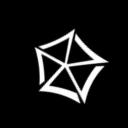}} & \raisebox{-\totalheight}{\includegraphics[width=1cm]{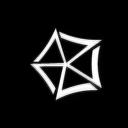}} & \raisebox{-\totalheight}{\includegraphics[width=1cm]{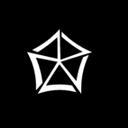}} & \raisebox{-\totalheight}{\includegraphics[width=1cm]{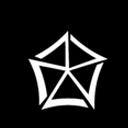}} & \raisebox{-\totalheight}{\includegraphics[width=1cm]{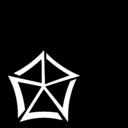}} & \raisebox{-\totalheight}{\includegraphics[width=1cm]{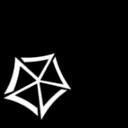}}    \\ 
      \raisebox{-\totalheight}{\includegraphics[width=1cm]{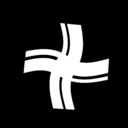}} & \raisebox{-\totalheight}{\includegraphics[width=1cm]{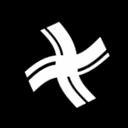}} & \raisebox{-\totalheight}{\includegraphics[width=1cm]{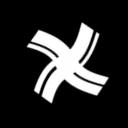}} & \raisebox{-\totalheight}{\includegraphics[width=1cm]{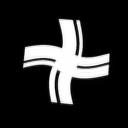}} & \raisebox{-\totalheight}{\includegraphics[width=1cm]{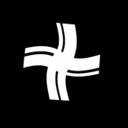}} & \raisebox{-\totalheight}{\includegraphics[width=1cm]{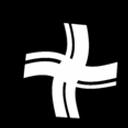}} & \raisebox{-\totalheight}{\includegraphics[width=1cm]{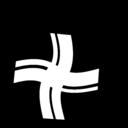}} & \raisebox{-\totalheight}{\includegraphics[width=1cm]{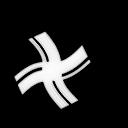}}    \\ 

  \bottomrule
      \raisebox{-\totalheight}{\includegraphics[width=1cm]{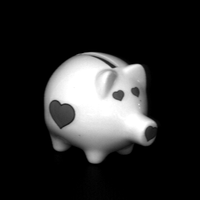}} & \raisebox{-\totalheight}{\includegraphics[width=1cm]{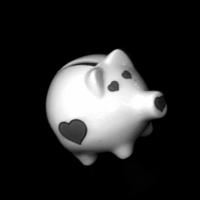}} & \raisebox{-\totalheight}{\includegraphics[width=1cm]{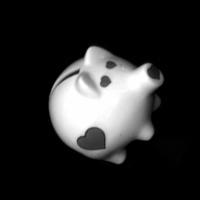}} & \raisebox{-\totalheight}{\includegraphics[width=1cm]{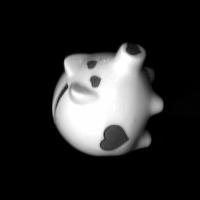}} & \raisebox{-\totalheight}{\includegraphics[width=1cm]{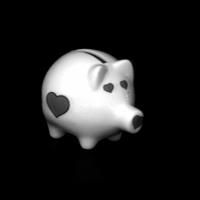}} & \raisebox{-\totalheight}{\includegraphics[width=1cm]{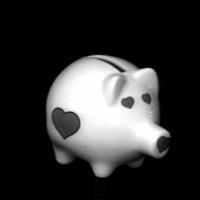}} & \raisebox{-\totalheight}{\includegraphics[width=1cm]{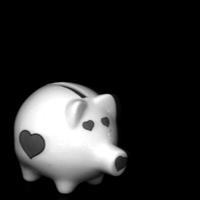}} & \raisebox{-\totalheight}{\includegraphics[width=1cm]{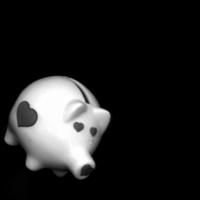}}    \\ 
      
      \raisebox{-\totalheight}{\includegraphics[width=1cm]{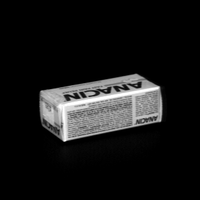}} & \raisebox{-\totalheight}{\includegraphics[width=1cm]{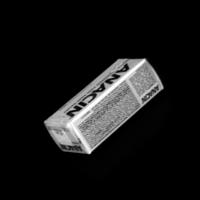}} & \raisebox{-\totalheight}{\includegraphics[width=1cm]{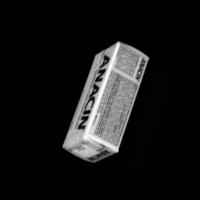}} & \raisebox{-\totalheight}{\includegraphics[width=1cm]{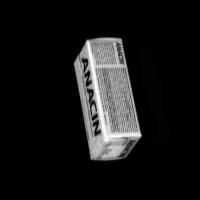}} & \raisebox{-\totalheight}{\includegraphics[width=1cm]{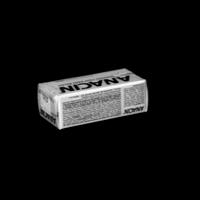}} & \raisebox{-\totalheight}{\includegraphics[width=1cm]{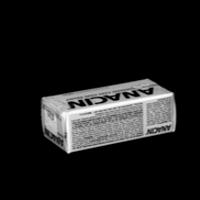}} & \raisebox{-\totalheight}{\includegraphics[width=1cm]{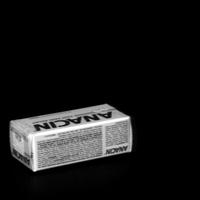}} & \raisebox{-\totalheight}{\includegraphics[width=1cm]{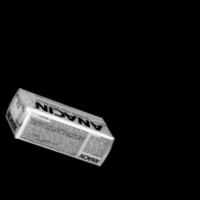}}    \\ 
            
          \raisebox{-\totalheight}{\includegraphics[width=1cm]{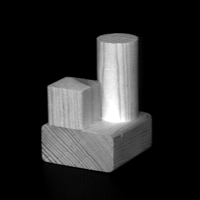}} & \raisebox{-\totalheight}{\includegraphics[width=1cm]{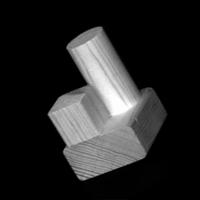}} & \raisebox{-\totalheight}{\includegraphics[width=1cm]{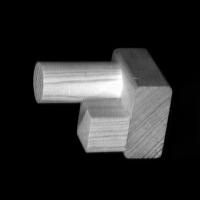}} & \raisebox{-\totalheight}{\includegraphics[width=1cm]{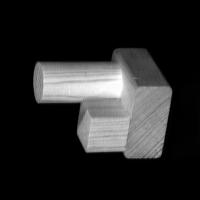}} & \raisebox{-\totalheight}{\includegraphics[width=1cm]{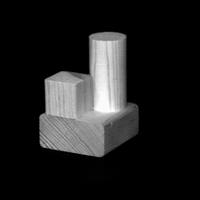}} & \raisebox{-\totalheight}{\includegraphics[width=1cm]{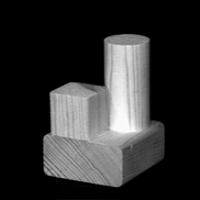}} & \raisebox{-\totalheight}{\includegraphics[width=1cm]{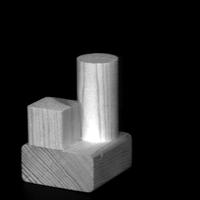}} & \raisebox{-\totalheight}{\includegraphics[width=1cm]{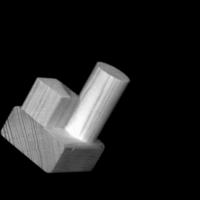}}    \\ 
  \bottomrule
 \end{tabular}
 }
  \caption{\footnotesize Generated images by TFT from the Kimia 
	, MPEG7 
	, and COIL datasets 
	. Images from different datasets are transformed using top-down transformers learned from the MNIST dataset. This is an inter dataset evaluation which demonstrates the effectiveness of our top-down model being generic and not tied to the training data.}
\label{table:GI}
\end{figure}

\subsection{Evaluation on non-MNIST Dataset}

\begin{table}[!htp]
\vspace{-1mm}
\centering
\caption{\footnotesize Mean squared pixel prediction error according to different affine transformations of DVAM and of our model on the notMNIST dataset.}
\vspace{2mm}
\scalebox{0.9}{
\begin{tabular}{l*{7}{c}r}
    \toprule
    Model & translation & rotation & scaling & combination\\
    \midrule
    DVAM & 0.091315 & 0.077126 & 0.056829 & 0.062878 \\
    DVAM+ & - & 0.08430 & - & - \\
    Ours & 0.001492 & 0.005311 & 0.004973 & 0.008571 \\
    \bottomrule
  \end{tabular}
}
\label{table:numerical_notmnist}
\end{table}

\begin{figure*}[!h]
  \centering
  \scalebox{1.0}{ 
  \begin{tabular}{m{1.8cm}m{1.8cm}m{1.8cm}m{1.8cm}m{1.8cm}m{1.8cm}m{1.8cm}}
    \toprule
    \centering
     & Original & Rotation (30\textdegree) & Rotation (-30\textdegree) & Scaling (x1.3) &  Scaling (x0.75) & Translation (up 30) \\ 
     \midrule
     FlowPCA \citep{gallagher2015happened} & \raisebox{-\totalheight}{\includegraphics[width=1.6cm]{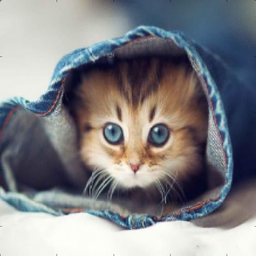}} & \raisebox{-\totalheight}{\includegraphics[width=1.6cm]{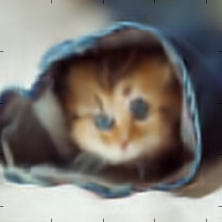}} & \raisebox{-\totalheight}{\includegraphics[width=1.6cm]{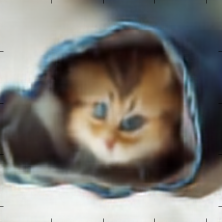}} & \raisebox{-\totalheight}{\includegraphics[width=1.6cm]{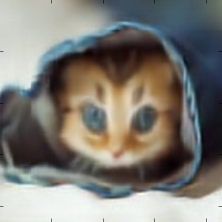}} &  \raisebox{-\totalheight}{\includegraphics[width=1.6cm]{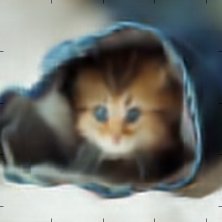}} & \raisebox{-\totalheight}{\includegraphics[width=1.6cm]{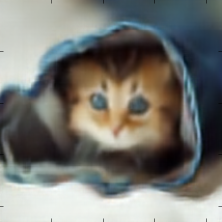}} \\ 
     \midrule
     Ours & \raisebox{-\totalheight}{\includegraphics[width=1.6cm]{cat/cat.png}} & \raisebox{-\totalheight}{\includegraphics[width=1.6cm]{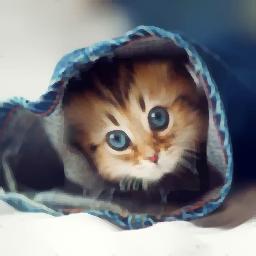}} & \raisebox{-\totalheight}{\includegraphics[width=1.6cm]{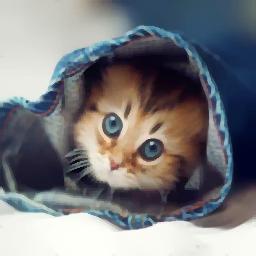}} & \raisebox{-\totalheight}{\includegraphics[width=1.6cm]{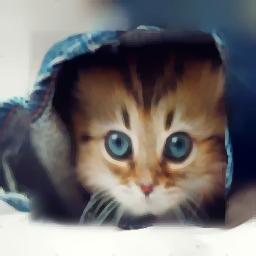}} & \raisebox{-\totalheight}{\includegraphics[width=1.6cm]{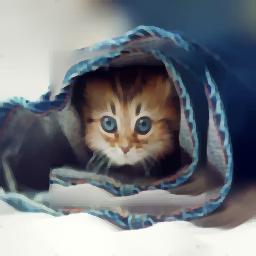}} & \raisebox{-\totalheight}{\includegraphics[width=1.6cm]{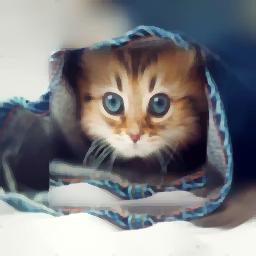}} \\ 
 \end{tabular}}
 \caption{Comparison of transformations on a natural image.}
 \vspace{-5mm}
\label{fig:natural_image}
\end{figure*}

We further test our model and the analogy network (DVAM) in \citep{reed2015deep} trained on the same MNIST dataset to investigate their inter dataset performance. In this experiment, we use our trained TFT based on the feature maps of the same traditional CNN, as discussed in section 5.1. To achieve similar visual and numerical effects to that of the MNIST dataset we normalize the notMNIST dataset using max norm. The results demonstrate that our networks have learned flow transformations of the CNN features that explicitly utilize top-down information and generalize well to new types of data. Numerical results are in Table \ref{table:numerical_notmnist} and visual reconstructions of transformed features are in Figure \ref{fig:comparison}.

Our model can be easily extended to feature maps of images with different sizes while \citep{reed2015deep} fails to do so. We apply the same learned TFT to CNN features of images from Kimia-99 \citep{kimia,belongie2002shape}, MPEG-7 Shape \citep{MPEG7} and COIL-20 \citep{Nene96columbiaobject} datasets. The generated images are illustrated in Figure \ref{table:GI}.

\vspace{-2mm}
\subsection{Evaluation on Natural Images}
\vspace{-0mm}
We also apply our learned TFT for natural images. Since the pre-trained network based on MNIST only accepts single channel images, we perform flow transformation to the colored images channel by channel. We compare our results with that of \citep{gallagher2015happened}, illustrated in Figure \ref{fig:natural_image}. We use a mask on the input CNN features when applying our method. 

\vspace{-0mm}
\subsection{Network Internal Data Augmentation}
\vspace{-0mm}

\begin{table}[!htp]
\vspace{-2mm}
\centering
\caption{\footnotesize ImageNet Validation Set Error (in \%).}
\vspace{2mm}
\scalebox{1.0}{
\begin{tabular}{lcccc} 
\multicolumn{1}{l}{Method}              &\shortstack{top-1} & \shortstack{top-5}
\\ \hline 
\scriptsize{VGG-16 \citep{Simonyan14c}}         & 24.8   & 7.5 \\
\scriptsize{VGG-16 fine-tuned via direct data aug.}         & 24.9   & 7.9 \\
\scriptsize{VGG-16 fine-tuned via TFT}     & 24.4   & 7.3 \\
\hline 
\end{tabular}
}
\label{table:numerical_imagenet}
\end{table}

As previously mentioned, the learned TFT can be used to further perform data augmentation inside CNN. In this experiment, we train our proposed TFT on CNN features out of the third pooling layer from the pre-trained VGG-16 \citep{Simonyan14c}. For this TFT, we set the depth of each transformation function as 2, the number of filters of the intermediate layer as 64, the filter size as 5x5, and the number of branhces $N$ as 16. We train the TFT using images synthesized from PASCAL VOC 2012 \citep{pascal-voc-2012}. In specific, we select 874 segmented images, each with single object approximately positioned in the center; we replace the background with the average color of images from the ImageNet dataset \citep{ILSVRC15} and generate rotation, translation and scaling transformations, similar to the settings in the previous experiments with MNIST. After training the model, we insert the TFT into the corresponding layer of the pre-trained VGG-16 and perform fine-tuning. We sample the top-down spatial control parameters  $\Theta = (\theta, s, t_x, t_y)$ uniformly from $[-30^o, 30^o] \times [0.8, 1.2] \times [-20, 20] \times [-20, 20]$ and accordingly achieve on-line data augmentation inside the networks. Our fine-tuning takes around 3 epochs for the network internal data augmentation. The results show marginal improvement as reported in Table \ref{table:numerical_imagenet}. 

We only use a few data for training TFT. And the original VGG-16 model already utilizes other direct data augmentation. In comparison, we perform fine-tuning using direct data augmentation (rotation, scaling and translation). As this creates unfilled holes and artifacts, it results in a slight increase to the top-1 and top-5 validation error. Using TFT provides improvement since TFT studies the intrinsic feature transformation (less artifacts) subject to the top-down transformation.

\vspace{-1mm}
\subsection{Image Style Transfer}
\vspace{-1mm}
\begin{figure}[!h]
 \vspace{-0mm}
  \centering
  \scalebox{0.7}{ 
  \begin{tabular}{cccc} 
    \centering
     & {\includegraphics[height=2.2cm]{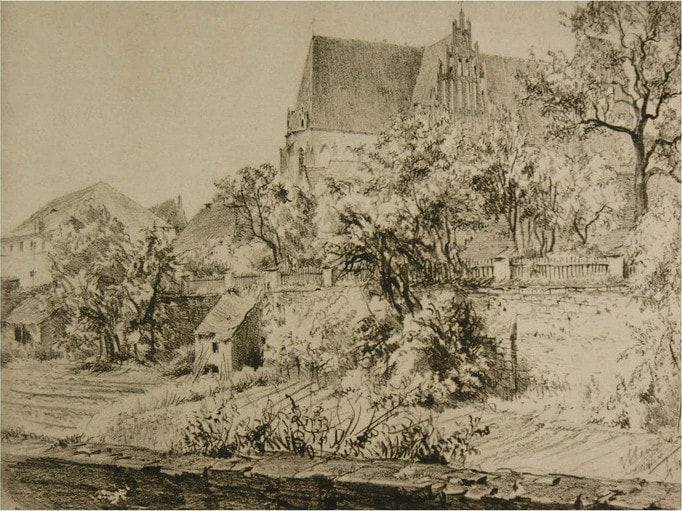}}  & {\includegraphics[height=2.2cm]{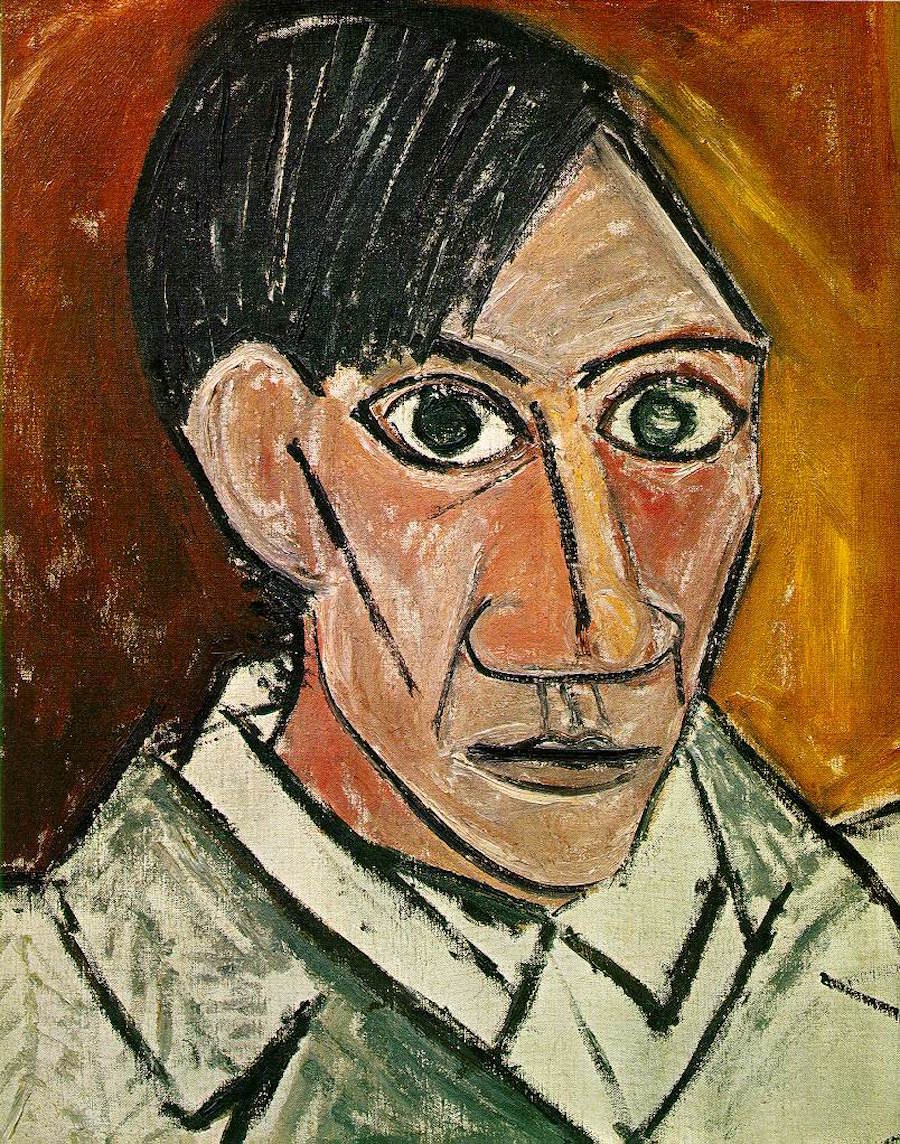}} &   \\ 
     & Style 1 & Style 2 \\
    \midrule
    \centering
     {\includegraphics[height=2.2cm]{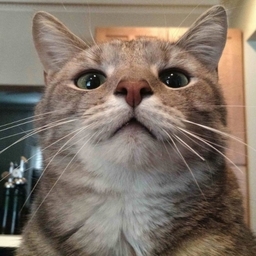}} &  {\includegraphics[width=2.2cm]{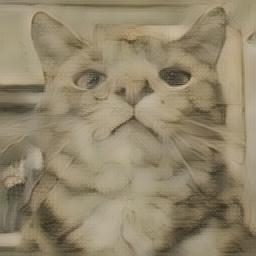}} 
     & {\includegraphics[width=2.2cm]{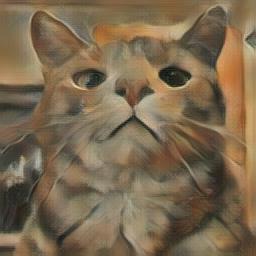}} & 
     {\includegraphics[width=2.2cm]{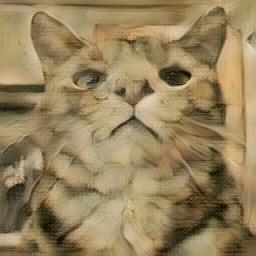}}  \\ 
     Input & Transfer: 1 & Transfer: 2 & Transfer: 1+2 \\
 \end{tabular}}
   \caption{Image style transfer experiments. 1+2 means a combined style of style 1 and 2.}
\label{fig:style}
\end{figure}
We also adapt TFT as stated in section 3.6 to the task of image style transfer. We select a few style images, and 200 images from MS-COCO \citep{DBLP:journals/corr/LinMBHPRDZ14} as content images and generate transformed images as 'ground truth' by the method in \citep{DBLP:journals/corr/LiFYWL017a}. We build TFT upon relu\_3\_1 of VGG-19 and leverage the pre-trained Decoder3 in \citep{DBLP:journals/corr/LiFYWL017a} to generate images from transformed feature maps. We set the depth of each transformation function as 5, the number of filters of the intermediate layers as 512.
An illustration is shown in Figure \ref{fig:style} for two individual styles and a combination of them.


\vspace{0mm}
\section{Conclusions}
\vspace{0mm}
We have developed top-down feature transformer (TFT) that learns a top-down generator by studying the internal transformations across CNN layers. The learned transformer is illustrated on both within and across datasets which demonstrates its clear advantage over models that are heavily learned through data-driven techniques. TFT points to a promising direction within the study of a CNN's internal representation and top-down processes.


{\small
\bibliographystyle{aaai}
\bibliography{aaai}
}

\end{document}